\documentclass[11pt,a4paper]{article}
\usepackage[latin9]{inputenc}
\usepackage{booktabs}
\usepackage{url}
\usepackage{svg}
\usepackage{amsmath}
\usepackage{graphicx}

\makeatletter

\providecommand{\tabularnewline}{\\}

\@ifundefined{date}{}{\date{}}
\usepackage[hyperref]{emnlp2018}
\usepackage{times}
\usepackage{latexsym}

\usepackage{url}

\makeatother
\aclfinalcopy 

\newcommand{\confname}{EMNLP 2018}

\title{Instructions for \confname{} Proceedings}

\author{
Shangbang Long$^1$,
\textbf{Cunchao Tu$^2$},
\textbf{Zhiyuan Liu}$^2$\thanks{Corresponding author.},
\textbf{Maosong Sun$^2$}\\
$^1$Peking University\\
$^2$Department of Computer Science and Technology\\
State Key Lab on Intelligent Technology and Systems\\
Institute for Artificial Intelligence, Tsinghua University, Beijing, China \\
{\small\tt longlongsb@pku.edu.cn, tucunchao@gmail.com, \{lzy,sms\}@tsinghua.edu.cn}\\
}

\begin{document}

\title{Automatic Judgment Prediction via Legal Reading Comprehension}
\maketitle
\begin{abstract}
Automatic judgment prediction aims to predict the judicial results
based on case materials. It has been studied for several decades mainly
by lawyers and judges, considered as a novel and prospective application
of artificial intelligence techniques in the legal field. Most existing
methods follow the text classification framework, which fails to model
the complex interactions among complementary case materials. To address
this issue, we formalize the task as \textbf{L}egal \textbf{R}eading
\textbf{C}omprehension according to the legal scenario. Following
the working protocol of human judges, LRC predicts the final judgment
results based on three types of information, including \emph{fact
description}, plaintiffs' \emph{pleas}, and \emph{law articles}. Moreover,
we propose a novel LRC model, \textbf{AutoJudge}, which captures the
complex semantic interactions among facts, pleas, and laws. In experiments,
we construct a real-world civil case dataset for LRC. Experimental
results on this dataset demonstrate that our model achieves significant
improvement over state-of-the-art models. We will publish all source codes and datasets of this work on \url{github.com}
for further research.
\end{abstract}

\section{Introduction}

Automatic judgment prediction is to train a machine judge to determine
whether a certain \emph{plea} in a given civil case would be supported
or rejected. In countries with\emph{ civil law system}, e.g. mainland
China, such process should be done with reference to related \emph{law
articles} and the \emph{fact description}, as is performed by a human
judge. The intuition comes from the fact that under \emph{civil law
system}, law articles act as principles for juridical judgments. Such
techniques would have a wide range of promising applications. 
On the one hand, legal consulting systems could provide better access
to high-quality legal resources in a low-cost way to legal outsiders,
who suffer from the complicated terminologies. On the other
hand, machine judge assistants for professionals would help improve
the efficiency of the judicial system. Besides, automated judgment
system can help in improving juridical equality and transparency.
From another perspective, there are currently 7 times much more civil
cases than criminal cases in mainland China, with annual rates of
increase of $10.8\%$ and $1.6\%$ respectively, making
judgment prediction in civil cases a promising application~\cite{ChinaLawBook}.

\begin{figure}
\centering 

\includegraphics[width=1\columnwidth]{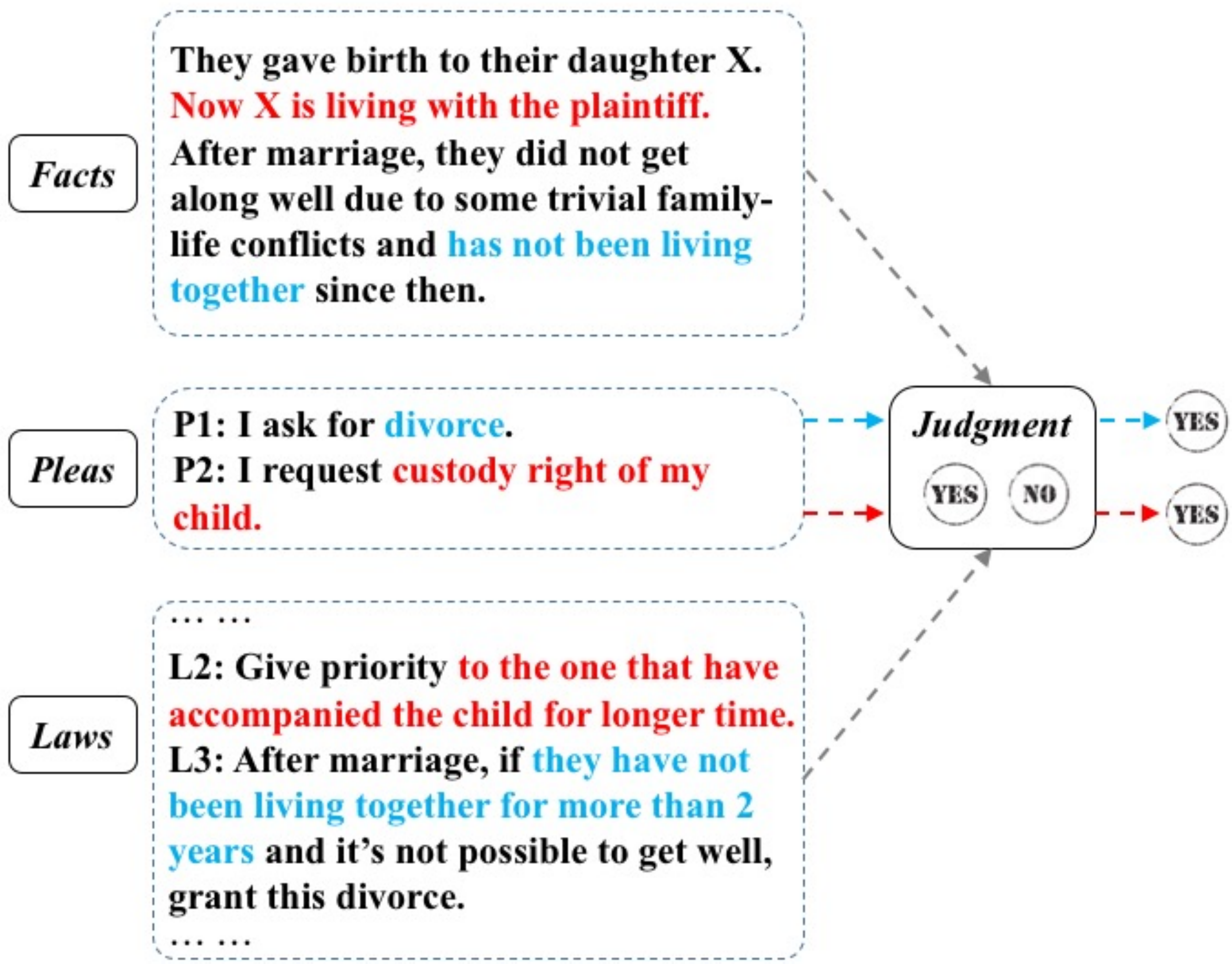}


\caption{An Example of LRC.}
\label{fig:lrc} 
\end{figure}

Previous works~\cite{aletras2016predicting,katz2017general,luo2017learning,Sulea2017Predicting}
formalize judgment prediction as the text classification task, regarding
either charge names or binary judgments, i.e., support or reject,
as the target classes. These works focus on the situation where only
one result is expected, e.g., the US Supreme Court's decisions~\cite{katz2017general},
and the charge name prediction for criminal cases~\cite{luo2017learning}.
Despite these recent efforts and their progress, automatic
judgment prediction in \emph{civil law system} is still confronted
with two main challenges:

\textbf{One-to-Many Relation between Case and Plea}. Every single
civil case may contain multiple pleas and the result of each plea
is co-determined by related law articles and specific aspects of the
involved case. For example, in divorce proceedings, judgment of \emph{alienation
of mutual affection} is the key factor for \emph{granting divorce}
but \emph{custody of children} depends on which side can provide better
an environment for children's growth as well as parents' financial
condition. Here, different pleas are independent.

\textbf{Heterogeneity of Input Triple}. Inputs to a judgment prediction
system consist of three heterogeneous yet complementary parts, i.e.,
\emph{fact description}, \emph{plaintiff's plea}, and \emph{related
law articles}. Concatenating them together and treating them simply
as a sequence of words as in previous works~\cite{katz2017general,aletras2016predicting}
would cause a great loss of information. This is the same in \emph{question-answering}
where the dual inputs, i.e., \emph{query} and \emph{passage}, should
be modeled separately.

Despite the introduction of the neural networks that can learn better
semantic representations of input text, it remains unsolved to incorporate
proper mechanisms to integrate the complementary triple of \emph{pleas},
\emph{fact descriptions}, and \emph{law articles} together.

Inspired by recent advances in question answering (QA) based reading
comprehension (RC) ~\cite{wang2017gated,cui2016attention,nguyen2016ms,rajpurkar2016squad}
, we propose the \textbf{Legal Reading Comprehension (LRC) } framework
for automatic judgment prediction. LRC incorporates the reading mechanism
for better modeling of the complementary inputs above-mentioned, as
is done by human judges when referring to legal materials in search
of supporting \emph{law articles}. Reading mechanism, by simulating
how human connects and integrates multiple text, has proven an effective module in RC tasks. We argue that applying the reading
mechanism in a proper way among the triplets can obtain a better
understanding and more informative representation of the original
text, and further improve 
performance
. To instantiate the
framework, we propose an end-to-end neural network model named
\textbf{AutoJudge}.

For experiments, we train and evaluate our models in the civil law
system of mainland China. We collect and construct a large-scale real-world
data set of $100,000$ case documents that the Supreme People's Court
of People's Republic of China has made publicly available
.
\emph{Fact description}, \emph{pleas}, and \emph{results} can be extracted
easily from these case documents with regular expressions, since the
original documents have special typographical characteristics indicating
the discourse structure. We also take into account law articles and
their corresponding juridical interpretations. We also implement and
evaluate previous methods on our dataset, which prove to be strong
baselines.

Our experiment results show significant improvements over previous
methods. Further experiments demonstrate that our model also achieves
considerable improvement over other off-the-shelf state-of-the-art
models under classification and question answering framework respectively.
Ablation tests carried out by taking off some components of our model
further prove its robustness and effectiveness.

To sum up, our contributions are as follows:

(1) We introduce reading mechanism and re-formalize judgment prediction
as Legal Reading Comprehension to better model the complementary inputs.

(2) We construct a real-world dataset for experiments, and plan to publish it for further research.

(3) Besides baselines from previous works, we also carry out comprehensive
experiments comparing different existing deep neural network methods
on our dataset. Supported by these experiments, improvements achieved
by LRC prove to be robust.

\section{Related Work}



\subsection{Judgment Prediction}

Automatic judgment prediction has been studied for decades. At the
very first stage of judgment prediction studies, researchers focus
on mathematical and statistical analysis of existing cases, without
any conclusions or methodologies on how to predict them~\cite{lauderdale2012supreme,segal1984predicting,keown1980mathematical,ulmer1963quantitative,nagel1963applying,kort1957predicting}.

Recent attempts consider judgment prediction under the text classification
framework. Most of these works extract efficient features from text
(e.g., N-grams)~\cite{Liu2017A,Sulea2017Predicting,aletras2016predicting,lin2012,liu2006exploring}
or case profiles (e.g., dates, terms, locations and types)~\cite{katz2017general}.
All these methods require a large amount of human effort to design
features or annotate cases. Besides, they also suffer from generalization
issue when applied to other scenarios.


Motivated by the successful application of deep neural networks, Luo
et al.~\cite{luo2017learning} introduce an attention-based
neural model to predict charges of criminal cases, and verify the
effectiveness of taking law articles into consideration. Nevertheless,
they still fall into the text classification framework and lack the
ability to handle multiple inputs with more complicated structures.

\subsection{Text Classification}

As the basis of previous judgment prediction works, typical text classification
task takes a single text content as input and predicts the category
it belongs to. Recent works usually employ neural networks to model
the internal structure of a single input~\cite{kim2014convolutional,Baharudin2010A,tang2015document,yang2016hierarchical}.

There also exists another thread of text classification called entailment
prediction. Methods proposed in~\cite{hu2014convolutional,Mitra2017Learning}
are intended for complementary inputs, but the mechanisms can be considered
as a simplified version of reading comprehension.

\subsection{Reading Comprehension}

Reading comprehension is a relevant task to model heterogeneous and
complementary inputs, where an \emph{answer} is predicted given two
channels of inputs, i.e. a textual \emph{passage} and a \emph{query}.
Considerable progress has been made~\cite{cui2016attention,dhingra2016gated,wang2017gated}.
These models employ various attention mechanism to model the interaction
between \emph{passage} and \emph{query}. Inspired by the advantage
of reading comprehension models on modeling multiple inputs, we apply
this idea into the legal area and propose legal reading comprehension
for judgment prediction.

\section{Legal Reading Comprehension}

\subsection{Conventional Reading Comprehension}

Conventional reading comprehension~\cite{Dureader,triviaqa,nguyen2016ms,rajpurkar2016squad}
usually considers reading comprehension as predicting the \emph{answer}
given a \emph{passage} and a \emph{query}, where the \emph{answer}
could be a single word, a text span of the original \emph{passage},
chosen from answer candidates, or generated by human annotators.

Generally, an instance in RC is represented as a triple $\langle p,q,a\rangle$,
where $p$, $q$ and $a$ correspond to $passage$, $query$ and $answer$
respectively. Given a triple $\langle p,q,a\rangle$, RC takes the
pair $\langle p,q\rangle$ as the input and employs attention-based
neural models to construct an efficient representation. Afterwards,
the representation is fed into the output layer to select or generate
an $answer$. 
\begin{figure*}
\centering 

\includegraphics[width=1.3\columnwidth]{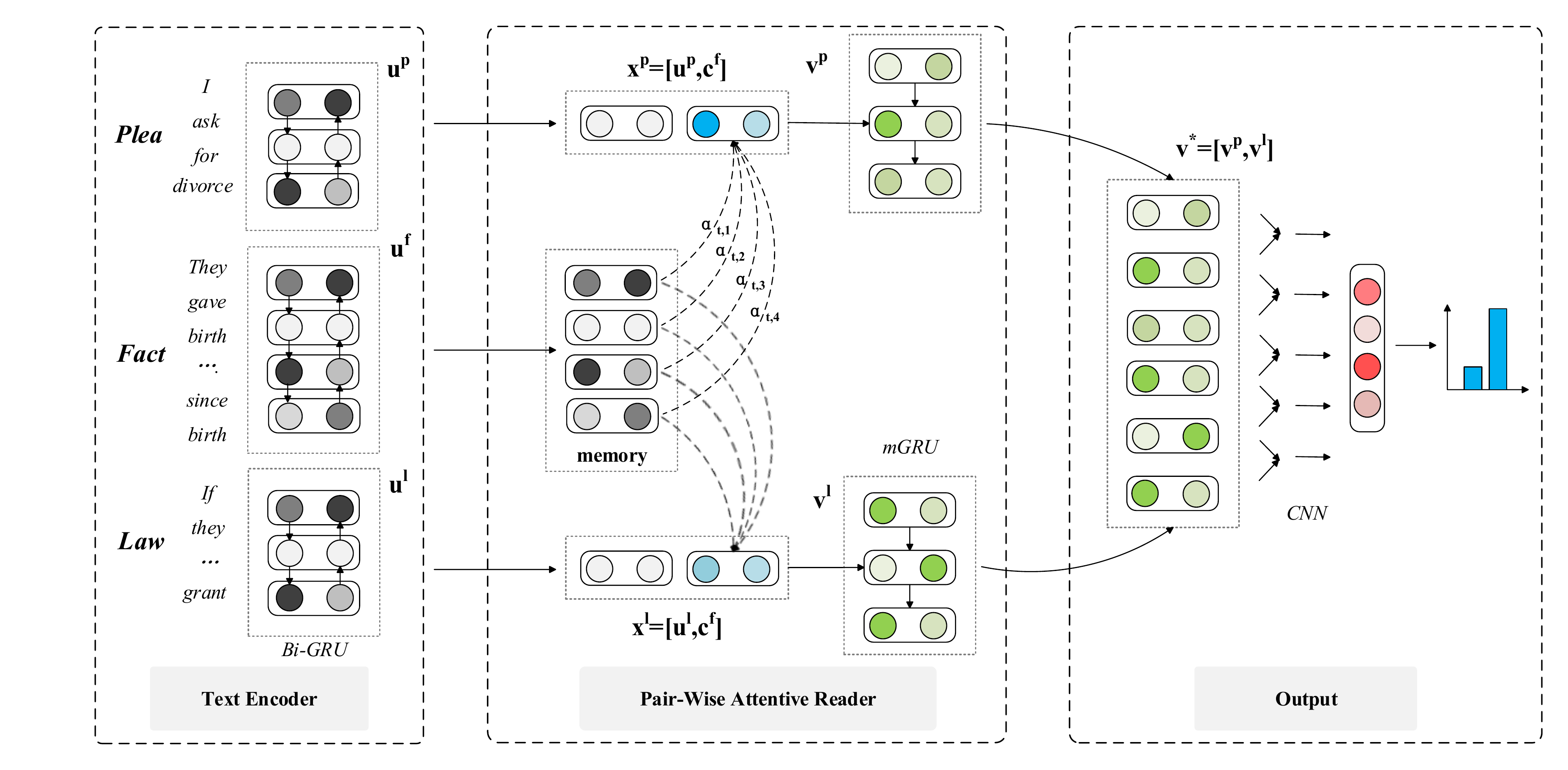}

\caption{An overview of AutoJudge.}
\label{fig:autojudge} 
\end{figure*}

\subsection{Legal Reading Comprehension}

Existing works usually formalize judgment prediction as a text classification
task and focus on extracting well-designed features of specific cases.
Such simplification ignores that the judgment of a case is determined
by its fact description and multiple pleas. Moreover, the final judgment
should act up to the legal provisions, especially in civil law systems.
Therefore, how to integrate the information (i.e., \emph{fact descriptions},
\emph{pleas}, and \emph{law articles}) in a reasonable way is critical
for judgment prediction.

Inspired by the successful application of RC, we propose a framework
of \textbf{L}egal \textbf{R}eading \textbf{C}omprehension(\textbf{LRC})
for judgment prediction in the legal area. As illustrated in Fig.~\ref{fig:lrc},
for each \emph{plea} in a given case, the prediction of judgment \emph{result}
is made based the \emph{fact description} and the potentially relevant
\emph{law articles}.

In a nutshell, LRC can be formalized as the following quadruplet task:
\begin{equation}
\small\langle f,\ p,\ l,\ r\rangle,
\end{equation}
where $f$ is the \emph{fact description, }$p$ is the \emph{plea},
$l$ is the \emph{law articles} and $r$ is the \emph{result}. Given
$\langle f,p,l\rangle$, LRC aims to predict the judgment result as

\begin{equation}
\small r=\mathop{\arg\max}_{r\in\{support,\ reject\}}P(r|f,p,l).
\end{equation}

The probability is calculated with respect to the interaction among
the triple $\langle f,\ p,\ l\rangle$, which will draw on the experience
of the interaction between $\langle passage,question\rangle$ pairs
in RC.

To summarize, LRC is innovative in the following aspects:

(1) While previous works fit the problem into text classification
framework, LRC re-formalizes the way to approach such problems. This
new framework provides the ability to deal with the heterogeneity
of the complementary inputs.

(2) Rather than employing conventional RC models to handle pair-wise
text information in the legal area, LRC takes the critical law articles
into consideration and models the facts, pleas, and law articles jointly
for judgment prediction, which is more suitable to simulate the human
mode of dealing with cases.

\section{Methods}

We propose a novel judgment prediction model \textbf{AutoJudge} to
instantiate the LRC framework. As shown in Fig.~\ref{fig:autojudge},
AutoJudge consists of three flexible modules, including a text encoder,
a pair-wise attentive reader, and an output module.

In the following parts, we give a detailed introduction to these three
modules.

\subsection{Text Encoder}

As illustrated in Fig.~\ref{fig:autojudge}, \textbf{Text Encoder}
aims to encode the word sequences of inputs into continuous representation
sequences.

Formally, consider a fact description $f=\{w_{t}^{f}\}_{t=1}^{m}$,
a plea $p=\{w_{t}^{p}\}_{t=1}^{n}$, and the relevant law articles
$l=\{w_{t}^{l}\}_{t=1}^{k}$, where $w_{t}$ denotes the $t$-th word
in the sequence and $m,n,k$ are the lengths of word sequences $f,p,l$
respectively. 

First, we convert the words to their respective word embeddings to
obtain $\mathbf{f}=\{\mathbf{w}_{t}^{f}\}_{t=1}^{m}$, $\mathbf{p}=\{\mathbf{w}_{t}^{p}\}_{t=1}^{n}$
and $\mathbf{l}=\{\mathbf{w}_{t}^{l}\}_{t=1}^{k}$, where $\textbf{w}\in {R}^{d}$.
Afterwards, we employ bi-directional GRU~\cite{Cho2014Learning,bahdanau2014neural,chung2014empirical}
to produce the encoded representation $\mathbf{u}$ of all words as
follows:

\begin{equation}
\small\begin{split}\mathbf{u}_{t}^{f} & =\text{BiGRU}_{F}(\mathbf{u}_{t-1}^{f},\mathbf{w}_{t}^{f}),\\
\mathbf{u}_{t}^{p} & =\text{BiGRU}_{P}(\mathbf{u}_{t-1}^{p},\mathbf{w}_{t}^{p}),\\
\mathbf{u}_{t}^{l} & =\text{BiGRU}_{L}(\mathbf{u}_{t-1}^{l},\mathbf{w}_{t}^{l}).
\end{split}
\label{eq:bigru}
\end{equation}
Note that, we adopt different bi-directional GRUs to encode fact descriptions,
pleas, and law articles respectively(denoted as $\text{BiGRU}_{F}$,
$\text{BiGRU}_{P}$, and $\text{BiGRU}_{L}$). With these text encoders,
$f$, $p$, and $l$ are converting into $\mathbf{u}^{f}=\{\mathbf{u}_{t}^{f}\}_{t=1}^{m}$,
$\mathbf{u}^{p}=\{\mathbf{u}_{t}^{p}\}_{t=1}^{n}$, and $\mathbf{u}^{l}=\{\mathbf{u}_{t}^{l}\}_{t=1}^{k}$.

\subsection{Pair-Wise Attentive Reader}

How to model the interactions among the input text is the most important
problem in reading comprehension. In AutoJudge, we employ a pair-wise
attentive reader to process $\langle\mathbf{u}^{f},\mathbf{u}^{p}\rangle$
and $\langle\mathbf{u}^{f},\mathbf{u}^{l}\rangle$ respectively. More
specifically, we propose to use pair-wise mutual attention mechanism
to capture the complex semantic interaction between text pairs, as
well as increasing the interpretability of AutoJudge.

\subsubsection{Pair-Wise Mutual Attention}

For each input pair $\langle\mathbf{u}^{f},\mathbf{u}^{p}\rangle$
or $\langle\mathbf{u}^{f},\mathbf{u}^{l}\rangle$, we employ pair-wise
mutual attention to select relevant information from fact descriptions
$\mathbf{u}^{f}$ and produce more informative representation sequences.

As a variant of the original attention mechanism~\cite{bahdanau2014neural},
we design the pair-wise mutual attention unit as a GRU with internal
memories denoted as \textbf{mGRU}.

Taking the representation sequence pair $\langle\mathbf{u}^{f},\mathbf{u}^{p}\rangle$
for instance, mGRU stores the fact sequence $\mathbf{u}^{f}$ into
its memories. For each timestamp $t\in[1,n]$, it selects relevant
fact information $\mathbf{c}_{t}^{f}$ from the memories as follows,
\begin{equation}
\small\mathbf{c}_{t}^{f}=\sum_{i=1}^{m}\alpha_{t,i}\mathbf{u}_{i}^{f}.
\end{equation}

Here, the weight $\alpha_{t,i}$ is the softmax value as 
\begin{equation}
\small\alpha_{t,i}=\frac{\exp(a_{t,i})}{\sum_{j=1}^{m}\exp(a_{t,j})}.\label{eq:attention}
\end{equation}

Note that, $a_{t,j}$ represents the relevance between $\mathbf{u}_{t}^{p}$
and $\mathbf{u}_{j}^{f}$. It is calculated as follows, 
\begin{equation}
\small a_{t,j}=\mathbf{V}^{T}\tanh(\mathbf{W}^{f}\mathbf{u}_{j}^{f}+\mathbf{W}^{p}\mathbf{u}_{t}^{p}+\mathbf{U}^{p}\mathbf{v}_{t-1}^{p}).
\end{equation}
Here, $\mathbf{v}_{t-1}^{p}$ is the last hidden state in the GRU,
which will be introduced in the following part. $\mathbf{V}$ is a
weight vector, and $\mathbf{W}^{f}$, $\mathbf{W}^{p}$, $\mathbf{U}^{p}$
are attention metrics of our proposed pair-wise attention mechanism.

\subsubsection{Reading Mechanism}

\label{sec:reading}

With the relevant fact information $\mathbf{c}_{t}^{f}$ and $\mathbf{u}_{t}^{p}$,
we get the $t$-th input of mGRU as 
\begin{equation}
\mathbf{x}_{t}^{p}=\mathbf{u}_{t}^{p}\oplus\mathbf{c}_{t}^{f},
\end{equation}
where $\oplus$ indicates the concatenation operation.

Then, we feed $\mathbf{x}_{t}^{p}$ into GRU to get more informative
representation sequence $\mathbf{v}^{p}=\{\mathbf{v}_{t}^{p}\}_{t=1}^{n}$
as follows,

\begin{equation}
\small\mathbf{v}_{t}^{p}=\text{GRU}(\mathbf{v}_{t-1}^{p},\mathbf{x}_{t}^{p}).
\end{equation}

For the input pair $\langle\mathbf{u}^{f},\mathbf{u}^{l}\rangle$,
we can get $\mathbf{v}^{l}=\{\mathbf{v}_{t}^{l}\}_{t=1}^{k}$ in the
same way. Therefore, we omit the implementation details Here.

Similar structures with attention mechanism are also applied in ~\cite{wang2017gated,rocktaschel2015reasoning,wang2015learning,bahdanau2014neural}
to obtain mutually aware representations in reading comprehension
models, which significantly improve the performance of this task.

\subsection{Output Layer}

Using text encoder and pair-wise attentive reader, the initial input
triple $\langle f,p,l\rangle$ has been converted into two sequences,
i.e., $\mathbf{v}^{p}=\{\mathbf{v}_{t}^{p}\}_{t=1}^{n}$ and $\mathbf{v}^{l}=\{\mathbf{v}_{t}^{l}\}_{t=1}^{k}$, where $\mathbf{v}_{t}^{l}$ is defined similarly to $\mathbf{v}_{t}^{p}$.
These sequences reserve complex semantic information about the pleas
and law articles, and filter out irrelevant information in fact descriptions.

With these two sequences, we concatenate $\mathbf{v}^{p}$ and $\mathbf{v}^{l}$
along the \emph{sequence length} dimension to generate the sequence
$\mathbf{v}^{\ast}=\{\mathbf{v}_{t}\}_{t=1}^{n+k}$. Since we have
employed several GRU layers to encode the sequential inputs, another
recurrent layer may be redundant. Therefore, we utilize a $1$-layer
CNN~\cite{kim2014convolutional} to capture the local structure and
generate the representation vector for the final prediction.

Assuming $y\in[0,1]$ is the predicted probability that the plea in
the case sample would be supported and $r\in\{0,1\}$ is the gold
standard, AutoJudge aims to minimize the cross-entropy as follows,
\begin{equation}
\small\mathcal{L}=-\frac{1}{N}\sum_{i=1}^{N}[r_{i}lny_{i}+(1-r_{i})ln(1-y_{i})],
\end{equation}
where $N$ is the number of training data. As all the calculation
in our model is differentiable, we employ Adam~\cite{kingma2014adam}
for optimization.

\section{Experiments}

To evaluate the proposed LRC framework and the AutoJudge model, we
carry out a series of experiments on the divorce proceedings, a typical
yet complex field of civil cases. Divorce proceedings often come with
several kinds of \emph{pleas}, e.g. \emph{seeking divorce}, \emph{custody
of children}, \emph{compensation}, and \emph{maintenance}, which focuses
on different aspects and thus makes it a challenge for judgment prediction.

\subsection{Dataset Construction for Evaluation}

\subsubsection{Data Collection}

Since none of the datasets from previous works have been published,
we decide to build a new one. We randomly collect $100,000$ cases
from China Judgments Online\footnote{\url{http://wenshu.court.gov.cn}},
among which $80,000$ cases are for training, $10,000$ each for validation
and testing. Among the original cases, $51\%$ are granted \emph{divorce}
and others not. There are $185,723$ valid \emph{pleas} in total,
with $52\%$ supported and $48\%$ rejected. Note that, if the \emph{divorce
plea} in a case is not granted, the other pleas of this case will
not be considered by the judge. Case materials are all natural language
sentences, with averagely $100.08$ tokens per \emph{fact description}
and $12.88$ per \emph{plea}. There are $62$ relevant \emph{law articles}
in total, each with $26.19$ tokens averagely. Note that the case documents include special typographical signals, making it easy to extract labeled data with regular expression.

\subsubsection{Data Pre-Processing }

We apply some rules with legal prior to preprocess the dataset according
to previous works~\cite{Liu2003Classification,liu2004case,Bian2005info},
which have proved effective in our experiments.

\textbf{Name Replacement}\footnote{We use regular expressions to extract \emph{names} and \emph{roles}
from the formatted case header. }\textbf{:} All names in case documents are replaced with marks indicating
their roles, instead of simply anonymizing them, e.g. \textless{}\emph{Plantiff}\textgreater{},
\textless{}\emph{Defendant}\textgreater{}, \textless{}\emph{Daughter\_x}\textgreater{}
and so on. Since ``\emph{all are equal before the law}''\footnote{Constitution of the People's Republic of China},
names should make no more difference than what role they take.

\textbf{Law Article Filtration} 
\textbf{:} Since most accessible divorce proceeding documents do
not contain ground-truth \emph{fine-grained} articles\footnote{\emph{Fine-grained} articles are in the \emph{Juridical Interpretations},
giving detailed explanation, while \emph{the} \emph{Marriage Law}
only covers some basic principles.}, we use an unsupervised method instead. First, we extract all the
articles from the law text with regular expression. Afterwards, we
select the most relevant 10 articles according to the fact descriptions
as follows. We obtain sentence representation with CBOW~\cite{mikolov2013efficient,mikolov2013distributed}
weighted by inverse document frequency, and calculate cosine distance
between cases and law articles. Word embeddings are pre-trained with
Chinese Wikipedia pages\footnote{\url{https://dumps.wikimedia.org/zhwiki/}}.
As the final step, we extract top $5$ relevant articles for each
sample respectively from the main marriage law articles and their
interpretations, which are equally important. We manually check the
extracted articles for $100$ cases to ensure that the extraction
quality is fairly good and acceptable.

The filtration process is automatic and fully unsupervised since the original
documents have no ground-truth labels for fine-grained law articles, and
coarse-grained law-articles only provide  limited information. We also
experiment with the ground-truth articles, but only a small fraction
of them has fine-grained ones, and they are usually not available
in real-world scenarios.


\subsection{Implementation Details}

We employ Jieba\footnote{\url{https://github.com/fxsjy/jieba}} for Chinese
word segmentation and keep the top $20,000$ frequent words. The word
embedding size is set to $128$ and the other low-frequency words
are replaced with the mark \emph{\textless{}UNK}\textgreater{}.
The hidden size of GRU is set to $128$ for each direction in Bi-GRU.
In the pair-wise attentive reader, the hidden state is set to $256$
for mGRu. In the CNN layer, filter windows are set to $1$, $3$,
$4$, and $5$ with each filter containing $200$ feature maps. We
add a dropout layer~\cite{srivastava2014dropout} after the CNN layer
with a dropout rate of $0.5$. We use Adam\cite{kingma2014adam} for
training and set learning rate to $0.0001$, $\beta_{1}$ to $0.9$
, $\beta_{2}$ to $0.999$, $\epsilon$ to $1e-8$, 
batch size to $64$. We employ \emph{precision}, \emph{recall}, \emph{F1}
and \emph{accuracy} for evaluation metrics. We repeat all the experiments
for $10$ times, and report the average results.

\subsection{Baselines}

For comparison, we adopt and re-implement three kinds of baselines
as follows:

\subparagraph{Lexical Features + SVM}

We implement an SVM with lexical features in accordance with previous
works~\cite{lin2012,liu2006exploring,aletras2016predicting,Liu2017A,Sulea2017Predicting}
and select the best feature set on the development set.

\subparagraph{Neural Text Classification Models}

We implement and fine-tune a series of neural text classifiers, including
attention-based method\cite{luo2017learning} and other methods we
deem important. CNN~\cite{kim2014convolutional} and GRU~\cite{Cho2014Learning,yang2016hierarchical}
take as input the concatenation of \emph{fact description} and \emph{plea}.
Similarly, \emph{CNN/GRU+law} refers to using the concatenation of
\emph{fact description}, \emph{plea }and\emph{ law articles }as inputs.

\subparagraph{RC Models}

We implement and train some off-the-shelf RC models, including
r-net\cite{wang2017gated} and AoA\cite{cui2016attention}, which
are the leading models on SQuAD leaderboard. In our initial experiments,
these models take \emph{fact description} as \emph{passage} and \emph{plea}
as \emph{query}. Further, \emph{Law articles} are added to the
\emph{fact description} as a part of the reading materials, which
is a simple way to consider them as well.

\subsection{Results and Analysis}

From Table~\ref{table:result}, we have the following observations:



\begin{table}
\centering 

{\small{}}%
\begin{tabular}{ccccc}
\toprule 
\\[-1.5em]
{\small{}Models } & {\small{}P } & {\small{}R } & {\small{}F1 } & {\small{}Acc. }\tabularnewline
\\[-1.5em]
\midrule 
\\[-1.5em]
{\small{}MaxFreq } & {\small{}$52.2$ } & {\small{}$100$ } & {\small{}$68.6$ } & {\small{}$52.2$}\tabularnewline
\\[-1.5em]
\midrule 
{\small{}SVM{*} } & {\small{}$57.8$ } & {\small{}$53.5$ } & {\small{}$55.6$ } & {\small{}$55.5$}\tabularnewline
\\[-1.5em]
\midrule 
\\[-1.5em]
{\small{}CNN } & {\small{}$76.1$ } & {\small{}$81.9$ } & {\small{}$79.0$ } & {\small{}$77.6$}\tabularnewline
\\[-1.5em]
{\small{}CNN+law } & {\small{}$74.4$ } & {\small{}$79.4$ } & {\small{}$77.0$ } & {\small{}$76.0$}\tabularnewline
\\[-1.5em]
{\small{}GRU } & {\small{}$79.2$ } & {\small{}$72.9$ } & {\small{}$76.1$ } & {\small{}$76.6$}\tabularnewline
\\[-1.5em]
{\small{}GRU+law } & {\small{}$78.2$ } & {\small{}$68.2$ } & {\small{}$72.8$ } & {\small{}$74.4$}\tabularnewline
\\[-1.5em]
{\small{}GRU+Attention{*} } & {\small{}$79.1$ } & {\small{}$80.7$ } & {\small{}$80.0$ } & {\small{}$79.1$}\tabularnewline
\\[-1.5em]
\midrule 
\\[-1.5em]
{\small{}AoA } & {\small{}$79.3$ } & {\small{}$78.9$ } & {\small{}$79.2$ } & {\small{}$78.3$}\tabularnewline
\\[-1.5em]
{\small{}AoA+law } & {\small{}$79.0$ } & {\small{}$79.2$ } & {\small{}$79.1$ } & {\small{}$78.3$}\tabularnewline
\\[-1.5em]
{\small{}r-net } & {\small{}$79.5$ } & {\small{}$78.7$ } & {\small{}$79.2$ } & {\small{}$78.4$}\tabularnewline
\\[-1.5em]
{\small{}r-net+law } & {\small{}$79.3$ } & {\small{}$78.8$ } & {\small{}$79.0$ } & {\small{}$78.3$}\tabularnewline
\\[-1.5em]
\midrule 
\\[-1.5em]
\textbf{\small{}AutoJudge}{\small{} } & \textbf{\small{}80.4}{\small{} } & \textbf{\small{}86.6}{\small{} } & \textbf{\small{}83.4}{\small{} } & \textbf{\small{}82.2}\tabularnewline
\\[-1.5em]
\bottomrule
\\[-1em]
\end{tabular}{\small{}\caption{Experimental results($\%$). P/R/F1 are reported for positive samples
and calculated as the mean score over 10-time experiments. Acc is
defined as the proportion of test samples classified correctly, equal
to micro-precision. MaxFreq refers to always predicting the most frequent
label, i.e. \emph{support} in our dataset. {*} indicates methods proposed
in previous works.}
\label{table:result} }{\small \par}
\end{table}

(1) AutoJudge consistently and significantly outperforms all the baselines,
including RC models and other neural text classification models, which
shows the effectiveness and robustness of our model.

(2) RC models achieve better performance than most text classification
models (excluding \emph{GRU+Attention}), which indicates that reading
mechanism is a better way to integrate information from
heterogeneous yet complementary inputs. 
On the contrary,
simply adding \emph{law articles} as a part of the reading materials
makes no difference in performance. Note that, \emph{GRU+Attention}
employ similar attention mechanism as RC does and takes additional
law articles into consideration, thus achieves comparable performance
with RC models.

(3) Comparing with conventional RC models, AutoJudge achieves significant
improvement with the consideration of additional law articles. It
reflects the difference between LRC and conventional RC models. We
re-formalize LRC in legal area to incorporate law articles via the
reading mechanism, which can enhance judgment prediction. Moreover,
CNN/GRU+law decrease the performance by simply concatenating original
text with law articles, while \emph{GRU+Attention/AutoJudge} increase
the performance by integrating law articles with attention mechanism.
It shows the importance and rationality of using attention mechanism
to capture the interaction between multiple inputs. 

The experiments support our hypothesis as proposed in the Introduction part that in civil cases, it's important to model the interactions among case materials. Reading mechanism can well perform the matching among them.

\begin{table}
\centering

{\small{}}%
\begin{tabular}{ccc}
\toprule 
\\[-1.5em]
{\small{}Models } & {\small{}F1 } & {\small{}Acc. }\tabularnewline
\\[-1.5em]
\midrule 
\\[-1.5em]
\textbf{\small{}AutoJudge}{\small{} } & {\small{}83.4} & {\small{}82.2}\tabularnewline
\\[-1.5em]
\midrule 
\\[-1.5em]
{\small{}w/o reading mechanism } & {\small{}78.9($\downarrow4.5$) } & {\small{}78.2($\downarrow4.0$)}\tabularnewline
\\[-1.5em]
\midrule 
\\[-1.5em]
{\small{}w/o law articles} & {\small{}79.6($\downarrow3.8$) } & {\small{}78.4($\downarrow3.8$)}\tabularnewline
\\[-1.5em]
\midrule 
\\[-1.5em]
{\small{}CNN$\rightarrow$LSTM } & {\small{}77.6($\downarrow5.8$) } & {\small{}77.7($\downarrow4.5$)}\tabularnewline
\\[-1.5em]
\midrule 
\\[-1.5em]
{\small{}w/o Pre-Processing } & {\small{}81.1($\downarrow2.3$) } & {\small{}80.3($\downarrow1.9$)}\tabularnewline
\\[-1.5em]
\midrule 
\\[-1.5em]
{\small{}w/o law article selection } & {\small{}80.6($\downarrow2.8$) } & {\small{}80.5($\downarrow1.7$)}\tabularnewline
\\[-1.5em]
\midrule 
\\[-1.5em]
{\small{}with GT law articles } & \textbf{\small{}85.1}{\small{}($\uparrow1.7$) } & \textbf{\small{}84.1}{\small{}($\uparrow1.9$)}\tabularnewline
\\[-1.5em]
\bottomrule
\\[-1.5em]
\end{tabular}{\small \par}

{\small{}\caption{Experimental results of ablation tests ($\%$). }
\label{table:abtest} }{\small \par}
\end{table}

\subsection{Ablation Test}

AutoJudge is characterized by the incorporation of pair-wise attentive
reader, law articles, and a CNN output layer, as well as some pre-processing
with legal prior. We design ablation tests respectively to evaluate
the effectiveness of these modules. When taken off the attention mechanism,
AutoJudge degrades into a GRU on which a CNN is stacked. When taken
off law articles, the CNN output layer only takes $\{v_{t}^{P}\}_{t=1}^{L_{P}}$
as input. Besides, our model is tested respectively without name-replacement
or unsupervised selection of law articles (i.e. passing the whole
law text). As mentioned above, we system use law articles extracted
with unsupervised method, so we also experiment with ground-truth
law articles.

Results are shown in Table~\ref{table:abtest}. We can infer that:

(1) The performance drops significantly after removing the attention
layer or excluding the law articles, which is consistent with the
comparison between AutoJudge and baselines. The result verifies that
both the reading mechanism and incorporation of law articles are important
and effective.

(2) After replacing CNN with an LSTM layer, performance drops as much
as $4.4\%$ in accuracy and $5.7\%$ in F1 score. The reason may be
the redundancy of RNNs. AutoJudge has employed several GRU layers
to encode text sequences. Another RNN layer may be useless to capture
sequential dependencies, while CNN can catch the local structure in
convolution windows.

(3) Motivated by existing rule-based works, we conduct data pre-processing
on cases, including name replacement and law article filtration. If
we remove the pre-processing operations, the performance drops considerably.
It demonstrates that applying the prior knowledge in legal filed would
benefit the understanding of legal cases.

\subparagraph{Performance Over Law Articles}

It's intuitive that the quality of the retrieved \emph{law articles}
would affect the final performance. As is shown in Table~\ref{table:abtest},
feeding the whole law text without filtration 
results in worse performance. However, when we train and evaluate
our model with ground truth articles, the performance is boosted by
nearly $2\%$ in both F1 and Acc. The performance improvement is quite
limited compared to that in previous work \cite{luo2017learning}
for the following reasons: (1) As mentioned above, most case documents
only contain coarse-grained articles, and only a small number of them
contain fine-grained ones, which has limited information in themselves.
(2) Unlike in criminal cases where the application of an article indicates
the corresponding crime, \emph{law articles} in civil cases work as
reference, and can be applied in both the cases of \emph{supports}
and \emph{rejects}. As law articles cut both ways for the judgment
result, this is one of the characteristics that distinguishes civil
cases from criminal ones. We also need to remember that, the performance
of $84.1\%$ in accuracy or $85.1\%$ in F1 score is unattainable
in real-world setting for automatic prediction where ground-truth
articles are not available.

\subparagraph{Reading Weighs More Than Correct Law Articles}


In the area of civil cases, the understanding of the case materials
and how they interact is a critical factor. The inclusion of law articles
is not enough. As is shown in Table \ref{table:abtest}, compared
to feeding the model with an un-selected set of law articles, taking
away the reading mechanism results in greater performance drop\footnote{$3.9\%$ vs. $1.7\%$ in Acc, and $4.4\%$ vs. $2.8\%$ in F1.}.
Therefore, the ability to read, understand and select relevant information
from the complex multi-sourced case materials is necessary. It's even
more important in real world since we don't have access to ground-truth
law articles to make predictions. 

\subsection{Case Study}

\subparagraph{Visualization of Positive Samples}

\begin{figure}
\centering

\includegraphics[width=0.8\columnwidth,height=1.1\columnwidth]{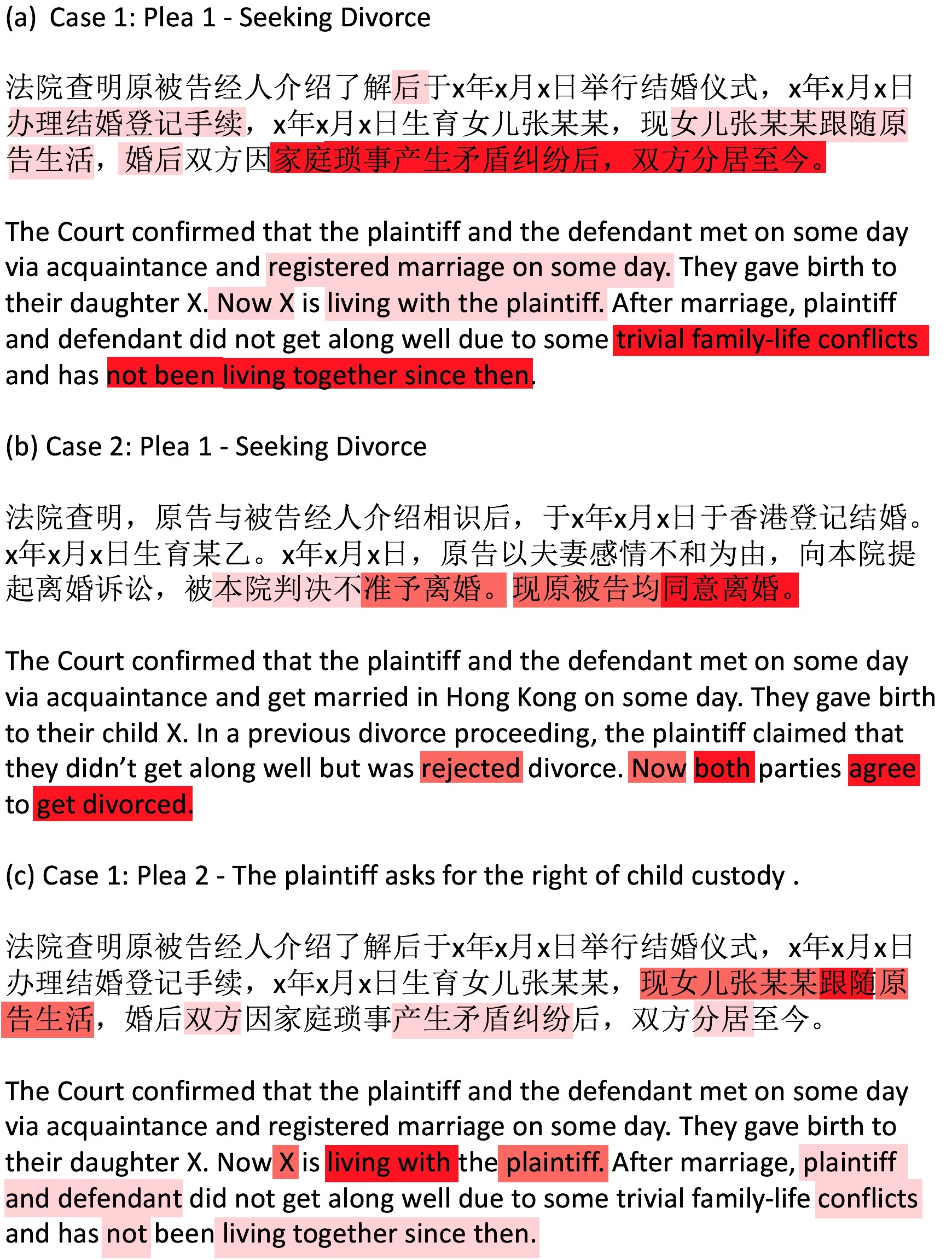}

\caption{Visualization of Attention Mechanism.}
\label{fig:case} 
\end{figure}

We visualize the heat maps of attention results\footnote{Examples given here are all drawn from the test set whose predictions
match the real judgment.}. As shown in Fig.~\ref{fig:case}, 
deeper background color represents larger
attention score.

The attention score is calculated with Eq. (\ref{eq:attention}).
We take the average of the resulting $n\times m$ attention matrix over the time dimension to obtain attention values for each word.

The visualization demonstrates that the attention mechanism can capture
relevant patterns and semantics in accordance with different \emph{pleas}
in different \emph{cases}.

\subparagraph{Failure Analysis}

As for the failed samples, the most common reason comes from the anonymity
issue, which is also shown in Fig.~\ref{fig:case}. As mentioned
above, we conduct name replacement. However, some critical elements
are also anonymized by the government, due to the privacy issue. These
elements are sometimes important to judgment prediction. For example,
determination of the key factor \emph{long-time separation} is relevant
to the explicit dates, which are anonymized. 

\section{Conclusion}

In this paper, we explore the task of predicting judgments of civil
cases. Comparing with conventional text classification framework,
we propose \textbf{L}egal \textbf{R}eading \textbf{C}omprehension
framework to handle multiple and complex textual inputs. Moreover,
we present a novel neural model, \textbf{AutoJudge}, to incorporate
law articles for judgment prediction. In experiments, we compare our
model on divorce proceedings with various state-of-the-art baselines of various frameworks. Experimental results
show that our model achieves considerable improvement than all the
baselines. Besides, visualization results also demonstrate the effectiveness
and interpretability of our proposed model.

In the future, we can explore the following directions: (1) 
Limited by the datasets, 
we can only verify our proposed model on divorce
proceedings. A more general and larger dataset will benefit the research
on judgment prediction. (2) Judicial decisions in some civil cases
are not always binary, but more diverse and flexible ones, e.g. compensation
amount. Thus, it is critical for judgment prediction to manage various
judgment forms.  

\bibliographystyle{acl_natbib_nourl}
\bibliography{emnlp2018}

\end{document}